\documentclass{article}

\usepackage[preprint]{nips_2018}
\usepackage[utf8]{inputenc} % allow utf-8 input
\usepackage[T1]{fontenc}    % use 8-bit T1 fonts
\usepackage{hyperref}       % hyperlinks
\usepackage{url}            % simple URL typesetting
\usepackage{booktabs}       % professional-quality tables
\usepackage{amsfonts}       % blackboard math symbols
\usepackage{amsmath}
\usepackage{amssymb}
\usepackage{nicefrac}       % compact symbols for 1/2, etc.
\usepackage{microtype}      % microtypography
\usepackage{times}
\usepackage{svg}
\usepackage{latexsym}
\usepackage{dirtytalk}
\usepackage{graphicx}
\usepackage{amssymb}
\usepackage{algorithm}
\usepackage{algpseudocode}
\usepackage[export]{adjustbox}
\usepackage{colortbl}

\DeclareMathOperator*{\argmin}{arg\,min}

\title{Siamese Capsule Networks}

\author{
  James O' Neill \\
  Department of Computer Science\\
  University of Liverpool\\
  Liverpool, L69 3BX \\
  \texttt{james.o-neill@liverpool.ac.uk} \\
}

\date{28-06-2017}

\begin{document}

\maketitle
% This is aimed for IJCAI or ICML, the other co-attention network paper is for ACL, best to forget NAACL, not enough time to prepare.
\begin{abstract}
Capsule Networks have shown encouraging results on \textit{defacto} benchmark computer vision datasets such as MNIST, CIFAR and smallNORB. Although, they are yet to be tested on tasks where (1) the entities detected inherently have more complex internal representations and (2) there are very few instances per class to learn from and (3) where point-wise classification is not suitable. Hence, this paper carries out experiments on face verification in both controlled and uncontrolled settings that together address these points. In doing so we introduce \textit{Siamese Capsule Networks}, a new variant that can be used for pairwise learning tasks. The model is trained using contrastive loss with $\ell_2$-normalized capsule encoded pose features. \iffalse Additionally, we use the probability of the final capsule layer activations as the individual dropout rate per capsule before passed to the final network layer that produces face embeddings.\fi We find that \textit{Siamese Capsule Networks} perform well against strong baselines on both pairwise learning datasets, yielding best results in the few-shot learning setting where image pairs in the test set contain unseen subjects.

\iffalse and replaces the nonlinear vector normalization activation function with a hyperbolic tangent function.\fi 

\iffalse
Thus, the main contribution in this paper the merging of Capsule Networks and siamese networks where an alternative loss function is used instead of the original capsule loss. Encodings of image pairs are made of combined pose matrices that represent the internal hierarchical representation of faces.
\fi
\iffalse Secondly, the application of Capsule Networks allows the model to learn if two images are similar but also what class they are from, while reducing intra-personal variations and maximize interpersonal differences.
\fi 
\iffalse
Secondly, we show few-shot face verification learning gives good performance on the both test datasets, reinforcing the notion that Capsule Networks need little data since they succinctly estimate internal hierarchical representations of entities.
\fi
\iffalse preserve information of object translations and rotations.\fi
\end{abstract}
 
\iffalse
The cortex is made of a hierarchy of cellular layers and within each layer there are \textit{minicolumns} which are organized group of neurons. The idea of capsules can be interpreted as a set of organized cells akin to the mini-columns hierarchical properties~\cite{kandel2000principles}. Adittionally, there are 10\% proximal excitatory neurons and 90\% distal excitatory neurons found in the cortex, in standard neural networks it is assumed proximal neurons are used as input to the activation function (i.e the cell in the cortex). \iffalse However, the distal synapses are active processing elements so that if a set of neurons are activated in close proximity at a particular point in time they are said to have a \textit{dendritic activation potential} that travel to the cell in a cortex and polarizes the cell having a large effect on it, not making it fire but depolarizing it. So now there is a large amount of synapses acting as a sort of coincidence detector. When learning in the brain occurs new connections can form and old ones drop out, this is similar to function \textit{DropConnect} plays in artificial neural networks. \fi
\fi
 \iffalse 
Such a routing scheme would rather route each part of input to the neurons that know how to deal with it, finding the best routing is equivalent to parsing and deciphering the image. 
\fi
 
\section{Introduction}
Convolutional Neural networks (CNNs) have been a mainstay model for a wide variety of tasks in computer vision. CNNs are effective at detecting local features in the receptive field, although the spatial relationship between features is lost when crude routing operations are performed to achieve translation invariance, as is the case with max and average pooling. Essentially, pooling results in viewpoint invariance so that small perturbations in the input do not effect the output. This leads to a significant loss of information about the internal properties of present entities (e.g location, orientation, shape and pose) in an image and relationships between them. The issue is usually combated by having large amounts of annotated data from a wide variety of viewpoints, albeit redundant and less efficient in many cases. As noted by ~{hinton1985shape}, from a psychology perspective of human shape perception, pooling does not account for the coordinate frames imposed on objects when performing mental rotation to identify handedness ~\cite{rock1973orientation,mcgee1979human,humphreys1983reference}. Hence, the scalar output activities from local kernel regions that summarize sets of local inputs are not sufficient for preserving reference frames that are used in human perception, since viewpoint information is discarded. Spatial transformer networks (STN) ~\cite{jaderberg2015spatial} have acknowledged the issue by using dynamic spatial transformations on feature mappings to enhance the geometric invariance of the model, although this approach addresses changes in viewpoint by learning to remove rotational and scale variance, as opposed to viewpoint variance being reflected in the model activations. Instead of addressing translation invariance using pooling operations, ~\cite{hinton2011transforming} have worked on achieving translation equivariance.\iffalse CNNs are not invariant to rotation and scale and more general affine transformations \fi The recently proposed Capsule Networks ~\cite{sabour2017dynamic,hinton2018matrix} have shown encouraging results to address these challenges. Thus far, Capsule Networks have only been tested on datasets that have (1) a relatively sufficient number of instances per class to learn from and (2) utilized on tasks in the standard classification setup. This paper extends Capsule Networks to the pairwise learning setting to learn relationships between whole entity encodings, while also demonstrating their ability to learn from little data that can perform zero-shot learning where instances from new classes arise during testing. The \textit{Siamese Capsule Network} is trained using a contrastive loss with $\ell_2$-normalized encoded features and demonstrated on two face verification tasks.

\section{Capsule Networks}\label{sec:capsule_nets}
~\cite{hinton2011transforming} first introduced the idea of using whole vectors to represent internal properties (referred to as instantiation parameters that include pose) of an entity with an associated activation probability where each capsule represents a single instance of an entity within in an image. This differs from the single scalar outputs in conventional neural networks where pooling is used as a crude routing operation over filters. Pooling performs sub-sampling so that neurons are invariant to viewpoint change, instead capsules look to preserve the information to achieve equivariance, akin to perceptual systems.
Hence, pooling is replaced with a dynamic routing scheme to send lower-level capsule (e.g nose, mouth, ears etc.) outputs as input to parent capsule (e.g face) that represent part-whole relationships to achieve translation equivariance and untangles the coordinate frame of an entity through linear transformations. The idea has its roots in computer graphics where images are rendered given an internal hierarchical representation, for this reason the brain is hypothesized to solve an \textit{inverse graphics} problem where given an image the cortex deconstructs it to its latent hierarchical properties. 
The original paper by ~\cite{sabour2017dynamic} describes a dynamic routing scheme that represent these internal representations as vectors given a group of designated neurons called capsules which consist of a pose vector $u \in \mathbb{R}^{d}$ and activation $\alpha \in [0,1]$. The architecture consists of two convolutional layers that are used as the initial input representations for the first capsule layer, which is then routed to a final class capsule layer. The initial convolutional layers allow learned knowledge from local feature representations to be reused and replicated in other parts of the receptive field. \iffalse from small kernels to be an initial point to then let the capsule layers estimate the pose and their relationship between such representations.\fi \iffalse In this we work we follow a similar scheme with the additional use of batch normalization.\fi The capsule inputs are determined using a \textit{Iterative Dynamic Routing} scheme. A transformation $W_{ij}$ is made to output vector $u_{i}$ of capsule $C^{L}_{i}$. The length of the vector $u_{i}$ represents the probability that this lower-level capsule detected a given object and the direction corresponds to the state of the object (e.g orientation, position or relationship to upper capsule). The output vector $u_i$ is transformed into a prediction vector $\hat{u}_{j|i}$, where $\hat{u}_{j|i} = W_{ij} u_i$. Then, $\hat{u}_{j|i}$ is weighted by a coupling coefficient $c_{ij}$ to obtain $s_{j} = \sum_{i} c_{ij}\hat{u}_{j|i}$, where coupling coefficients for each capsule $\sum_j c_{ij}=1$ and $c_{ij}$ is got by log prior probabilities $b_{ij}$ from a sigmoid function, followed by the $\mathtt{softmax}$, $c_{ij}= e^{b_{ij}}/\sum_{k} e^{b_{ik}}$. If $\hat{u}^{L}_{j|i}$ has high scalar magnitude when multiplied by $u^{L+1}_{j}$ then the coupling coefficient $c_{ij}$ is increased and the remaining potential parent capsules coupling coefficients are decreased. \textit{Routing By Agreement} is then performed using coincidence filtering to find tight clusters of nearby predictions. \iffalse from all $s_{j}$ predictions for capsule $C^{L+1}$.\fi The entities output vector length is represented as the probability of an entity being present by using the nonlinear normalization shown in Equation \ref{eq:squash} where vote $v_j$ is the output from total input $s_j$, which is then used to compute the agreement $a_{ij} = v_{j}\hat{u}_{j|i}$ that is added to the log prior $b_{ij}$.  

\iffalse
We want the length of the output vector of a capsule $s_j$ to represent the probability that the entity represented by the capsule is present in the current input. Hence, a nonlinear function (necessary for learning also) normalizes the vector lengths in range [0,1]. Here vote $v_j$ is  the vector output of capsule $j$ and ${\mathbf{s}}_j$ is its total input.
\fi

\begin{equation}\label{eq:squash}
{\bf v}_j = \frac{||{\bf s}_j||^2}{1+||{\bf s}_j||^2} \frac{{\bf s}_j}{||{\bf s}_j||}
\end{equation}

The capsule is assigned a high log-likelihood if densely connected clusters of predictions are found from a subset of $s$. The centroid of the dense cluster is output as the entities generalized pose. This coincidence filtering step can also be achieved by traditional outlier detection methods such as Random sample consensus (RANSAC) ~\cite{fischler1987random} and classical Hough Transforms ~\cite{ballard1987generalizing} for finding subsets of the feature space with high agreement. Although, the motivation for using the vector normalization of the instantiation parameters is to force the network to preserve orientation. Lastly, a reconstruction loss on the images was used for regularization which constrains th capsules to learn properties that can better encode the entities. In this paper, we do not use such regularization scheme by autoencoding pairs of input images, instead we use a variant of dropout.

\paragraph{Extensions Of Capsule Networks}
~\cite{hinton2018matrix} recently describe matrix capsules that perform routing by agreement using the expectation maximization (EM) algorithm, motivated by computer graphics where pose matrices are used to define rotations and translations of objects to account for viewpoint changes. Each parent capsule is considered a Gaussian and the pose matrix of each child capsule are considered data samples of the Gaussian.
A given layer $L$ contains a set of capsules $C^{L}$ such that $\forall C^{\ell}   \exists \: \{\mathcal{M}^{\ell}, \alpha^{\ell}\} \: \: \in C^{L}$ where pose matrix $\mathcal{M}^{\ell} \in \mathbb{R}^{n \times n}$ ($n=4$) and activation $\alpha^{\ell} \in [0, 1]$ are the outputs. A vote is made $V_{ij} = \mathcal{M}_{i}W_{ij}$ for the pose matrix of $C^{L+1}_{j}$ where $W_{ij} \in \mathbb{R}^{n\times n}$ is a learned viewpoint invariant transformation matrix from capsule $C^{L}_{i} \to C^{L+1}_{j}$.
\iffalse Each $c_{ij}$ is a non-negative scalar such that $\sum_i c_{ij}=1$ no need to mention already did for the previous part\fi 
EM determines the activation of $C^{L+1}_{j}$ as $a_{j} = \sigma\big(\lambda(\beta_{a} - \beta_{u}\sum_{i}r_{ij} - \sum_{h}cost^{h}_{j})\big)$ where the $cost^{h}_j$ is the negative log-probability density weighted by the assignment probabilities $r_{ij}$, $-\beta_{u}$ is the negative log probability density per pose matrix computed to describe $C^{L+1}_{j}$. If $C^{L+1}_{j}$ is activated $-\beta_{a}$ is the cost for describing ($\mu_j$, $\sigma_j^{2}$) from lower-level pose data samples along with $r_{ij}$ and $\lambda$ is the inverse temperature so as the assignment probability becomes higher the slope of the sigmoid curve becomes steeper (represents the presence of an entity instead of the nonlinear vector normalization seen in Equation \ref{eq:squash}). The network uses 1 standard convolutional layer, a primary capsule layer, 2 intermediate capsule convolutional layer, followed by the final class capsule layer. The matrix capsule network significantly outperformed CNNs on the SmallNORB dataset.
\iffalse
The paper explains the choice between computing lower-level capsules pose when they are not used to activate a given parent capsule and the computational cost in computing mean and variance for that capsule if activated. The difference in cost between these choices is passed to a sigmoid to compute the upper layer activation probability.
\fi

\iffalse 
 This update is used to route capsule outputs to upper capsules that take a cluster of similar cluster.
 Instead, a vote is taken for pose matrices in upper layers from capsules in the previous layer where each vote takes a weighted assignment coefficient that are updated with the expectation-maximization (EM) algorithm for three iterations per pairs of neighboring layers.
 The cost associated with activating a capsule given all poses is the fraction of lower capsules that activate a capsule in the next layer. The cost for describing the data samples is given by the negative log probability density of that data samples vote on the Gaussian distribution fitted by a given higher-level capsule is gets assigned to.
\fi

%%%%%%%%%%%%%%%
\iffalse
In this paper, 
\textit{and the fact that it is active and then pay additional costs, pro-rated by the assignment probabilities, for describing the discrepancies between the lower-level means and the values predicted for them when the mean of the higher-level capsule is used to predict them via the inverse of the transformation matrix.} 
\fi
%%%%%%%%%%%%%%%

~\cite{lalonde2018capsules} introduce SegCaps which uses a locally connected dynamic routing scheme to reduce the number of parameters while using deconvolutional capsules to compensate for the loss of global information, showing best performance for segmenting pathological lungs from low dose CT scans. The model obtained a 39\% and 95\% reduction in parameters over baseline architectures while outperforming both.\iffalse https://openreview.net/pdf?id=ByVJlqR8G \fi ~\cite{bahadori2018spectral} introduced Spectral Capsule Networks demonstrated on medical diagnosis. The method shows faster convergence over the EM algorithm used with pose vectors. Spatial coincidence filters align extracted features on a 1-d linear subspace. The architecture consists of a 1d convolution followed by 3 residual layers with dilation. Residual blocks $R$ are used as nonlinear transformations for the pose and activation of the first primary capsule instead of the linear transformation that accounts for rotations in CV, since deformations made in healthcare imaging are not fully understood. The weighted votes are obtained as $s_{j,i} = \alpha_{i} R_{j}(u_i) \quad \forall i$ where $S_j$ is a matrix of concatenated votes that are then decomposed using SVD, where the first singular value dimension $\tilde{s}_{1}$ is used to capture most of the variance between votes, thus the activation $a_j$ activation is computed as $\sigma\big(\eta(s^{2}_1/\sum_{k} s_{k}^{2} - b)\big)$ where $s^{2}_1/\sum_{k} s_{k}^{2}$ is the ratio of all variance explained for all right singular vectors in $V$, $b$ is optimized and $\eta$ is decreased during training. The model is trained by maximizing the log-likelihood showing better performance than the spread loss used with matrix capsules and mitigates the problem of capsules becoming dormant.
%Xi et al. ~\cite{xi2017capsule} test capsule networks on complex data, 

~\cite{wang2018optimization} formalize the capsule routing strategy as an optimization of a clustering loss and a KL regularization term between the coupling coefficient distribution and its past states. The proposed objective function follows as $\min_{C,S}\{\mathcal{L}(C,S):= - \sum_{i}\sum_{j}c_{ij} \langle o_{j|i}, s_{j} \rangle + \alpha \sum_{i}\sum_{j}c_{ij} \log c_{ij}\}$ where $o_{j|i} = T_{ij}\mu_{i}/||T_{ij}||_{\mathcal{F}}$ and $||T_{ij}||_{\mathcal{F}}$ is the Frobenious norm of $T_{ij}$. This routing scheme shows significant benefit over the original routing scheme by ~\cite{sabour2017dynamic} as the number of routing iterations increase.
Evidently, there has been a surge of interest within the research community. This paper looks to contribute primarily to how Capsule Network can be used for comparing images, aligning entities across images and describing a method for measuring similarity between final layer capsules such that inter-class variations are maximized and intra-class variations are minimized. First, we briefly describe state of the art work in face verification that have used Siamese Networks. 

% motivated by crowding - pelli2004crowding: Crowding is unlike ordinary masking: Distinguishing feature integration from detection --- binding hinton1981shape

%%%%%%%%%%%%%%%%%% ------- SIAMESE NETWORK --- %%%%%%%%%%%%%%%%%%%%%%%%%%%%

\section{Siamese Networks For Face Verification}\label{sec:siamese_net}
Siamese Networks (SNs) are neural networks that learn relationships between encoded representations of instance pairs that lie on low dimensional manifold, where a chosen distance function $d_{\omega}$ is used to find the similarity in output space. Below we briefly describe state of the art convolutional SN's that have been used for face verification and face recognition.

~\cite{sun2014deep} presented a joint identification-verification approach for learning face verification with a contrastive loss and face recognition using cross-entropy loss. To balance loss signals for both identification and verification, they investigate the effects of varying weights controlled by $\lambda$ on the intra-personal and inter-personal variations, where $\lambda=0$ leaves only the face recognition loss and $\lambda\to \infty$ leaves the face verification loss. Optimal results are found when $\lambda=0.05$ intra personal variation is maximized while both class are distinguished. 

~\cite{wen2016discriminative} propose a center loss function to improve discriminative feature learning in face recognition. The center loss function proposed aims to improve the discriminability between feature representations by minimizing the intra-class variation while keeping features from different classes separable. The center loss is given as $\mathcal{L} = -\sum_{i=1}^{m} \log (e^{z})/(\sum_{j=1}^{n}e^{z}) + \lambda_{2} \sum_{m}{i=1}||x_{i} - c_{y_{i}}||^{2}_{2}$ where $z = W^{T}_{j}x_{i} + b_{j}$. The $c_{y_{i}}$ is the centroid of feature representations pertaining to the $i^{th}$ class. This penalizes the distance between class centers and minimizes the intra-class variation while the $\mathtt{softmax}$ keeps the inter-class features separable. The centroids are computed during stochastic gradient descent as full batch updates would not be feasible for large networks.

~\cite{liu2017sphereface} proposed \textit{Sphereface}, a hypersphere embedding that uses an angular softmax loss that constrains disrimination on a hypersphere manifold, motivated by the prior that faces lie on a manifold. The model achieves 99.22 \% on the LFW dataset, and competitive results on Youtube Face (YTF) and MegaFace.
~\cite{sankaranarayanan2016triplet} proposed a triplet similarity embedding for face verification using a triple loss $\argmin_{W} = \sum_{\alpha,p,n\in T} \max(0, \alpha+ \alpha^{T}W^{T}W(n -p))$ where for $T$ triplet sets lies an anchor class $\alpha$, positive class $p$ and negative class $n$, a projection matrix $W$, (performed PCA to obtain $W_{0}$) is minimized with the constraint that $W_{a}^{T}W_{p} > W_{a}^{T}W_{n}$. The update rule is given as $W_{t+1} = W_{t} - \eta W_{t} (\alpha(n-p)^{T} + (n-p)\alpha^{T})$. \iffalse Instead of using a linear transformation using symmetric and positive semi-definite matrix $M= W^{T}W$ in Mahalaobonis Distance Metric Learning where $d_{M} = \sqrt{(x_{i} - x_{j})^{T}M(x_{i} - x_{j})}$,\textbf{here}\fi
~\cite{hu2014discriminative} use deep metric learning for face verification with loss $\argmin_{f} J = \frac{1}{2}\sum_{i,j} g\big(1 - \ell_{i,j}(\tau - d^{2}_{f}(x_{i},x_{j})))\big)+ \frac{\lambda}{2}(\sum_{m=1}^{M}\big(|\theta^{(m)}|^{2}_{F}|\big)$ where $g(z) = \log(1+e^{\beta z})/\beta$, $\beta$ controls the slope steepness of the logistic function, $||A||_{\mathcal{F}}$ is the frobenius norm of $A$ and $\lambda$ is a regularization parameter. Hence, the loss function is made up of a logistic loss and regularization on parameters $\theta=[W,b]$. Best results are obtained using a combination of SIFT descriptors, dense SIFT and local binary patterns (LBP), obtaining 90.68\% (+/- 1.41) accuracy on the LFW dataset.

~\cite{ranjan2017l2} used an $\ell_2$-constraint on the softmax loss for face verification so that the encoded face features lie on the ambit of a hypersphere, showing good improvements in performance. This work too uses an $\ell_2$-constraint on capsule encoded face embeddings.\iffalse, however this is for the purposes of retrieving the accuracy from the margin set from the contrastive loss. \fi \iffalse
~\cite{hoffer2015deep} have proposed a triplet network that takes three input images, provides $l_2$ distance between the output of three separate but identical CNNs ($f(x_+,x),(x,x_-)$ where $x$ and $x_+$ are the same class and $x_-$ are not) demonstrated on digit recognition and animal/vehicle image classification.
\fi
FaceNet ~\cite{schroff2015facenet} too uses a triplet network that combines the Inception network ~\cite{szegedy2015going} and a 8-layer convolutional model ~\cite{zeiler2014visualizing} which learns to align face patches during training to perform face verification, recognition and clustering. The method trains the network on triplets of increasing difficulty using a negative example mining technique. Similarly, we consider a Siamese Inception Network for the tasks as one of a few comparisons to SCNs.
%Parkhi ~\cite{parkhi2015deep}

The most relevant and notable use of Siamese Networks for face verification is the \textit{DeepFace} network, introduced by \cite{taigman2014deepface}. The performance obtained was on par with human level performance on the \textit{Faces in the Wild} (LFW) dataset and significantly outperformed previous methods\iffalse improved over previous results by a 27\% percentage point increase\fi. However, it is worth noting this model is trained on a large dataset from Facebook (SFC), therefore the model can be considered to be performing transfer learning before evaluation. The model also carries out some manual steps for detecting, aligning and cropping faces from the images. For detecting and aligning the face a 3D model is used. The images are normalized to avoid any differences in illumination values, before creating a 3D model which is created by first identifying 6 fiducial points in the image using a Support Vector Regressor from a LBP histogram image descriptor. Once the faces are cropped based on these points, a further 67 fiducial point are identified for 3D mesh model, followed by a piecewise affine transformation for each section of the image. The cropped image is then passed to 3 CNN layers with an initial max-pooling layer followed two fully-connected layers. Similar to Capsule Networks, the authors refrain from using max pooling at each layer due to information loss. In contrast to this work, the only preprocessing steps for the proposed SCNs consist of pixel normalization and a reszing of the image. 

The above work all achieve comparable state of the art results for face verification using either a single CNN or a combination of various CNNs, some of which are pretrained on large related datasets. In contrast, this work looks to use a smaller Capsule Network that is more efficient, requires little preprocessing steps (i.e only a resizing of the image and normalization of input features, no aligning, cropping etc.) and can learn from relatively less data.

\section{Siamese Capsule Network}\label{sec:method}
The Capsule Network for face verification is intended to identify enocded part-whole relationships of facial features and their pose that in turn leads to an improved similarity measure by aligning capsule features across paired images. The architecture consists of a 5-hidden layer (includes 2 capsule layers) network with tied weights (since both inputs are from the same domain). The $1^{st}$ layer is a convolutional filter with a stride of 3 and 256 channels with kernels $\kappa^{1}_{i} \in \mathbb{R}^{9 \times 9} \; \forall i$  over the image pairs $\langle x_1,x_2 \rangle \in \mathbb{R}^{100 \times 100}$ , resulting in 20, 992 parameters. The $2^{nd}$ layer is the primary capsule layer that takes $\kappa^{(1)}$ and outputs $\kappa^{(2)} \in \mathbb{R}^{31 \times 31}$ matrix for 32 capsules, leading to $5.309\times 10{^6}$ parameters (663, 552 weights and 32 biases for each of 8 capsules). The $3^{rd}$ layer is the face capsule layer, representing the routing of various properties of facial features, consisting of $5.90 \times 10^{6}$ parameters. This layer is then passed to a single fully connected layer by concatenating the pose vectors $\mathcal{M}^{L}_{\cap} = \cap_{i=1}^{|C^{L}|}$ as input, while the sigmoid functions control the dropout rate for each capsule during training. The nonlinear vector normalization shown in Equation \ref{eq:squash} is replaced with a tanh function $\tanh(.)$ which we found in initial testing to produce better results. Euclidean distance, Manhattan distance and cosine similarity are considered as measures between the capsule image encodings. The aforementioned SCN architecture describes the setup for the AT\&T dataset. For the LFW dataset, 6 routing iterations are used and 4 for AT\&T.

\iffalse
The third layer is the primary capsule layer that takes the 256 output kernels where each capsule multiplies this by 8 convolutions of size $9 \times 9 \times 256$ to the input from layer 1 of size $84\times 20 \times 512$. This outputs a $\times \times$ for each capsule, so for all capsules the output is $6\times 6 \times 8 \times 32$ which leads to $5.309\times 10{^6}$ parameters. 
\fi

\iffalse
% Hausdorf distance 
\begin{equation}
	d_{H}(h^{(\ell)}_{1}, h^{(\ell)}_{2}) = \max\Big\{\sup_{h^{(\ell)}_{1} \in H^{\ell}} \inf_{h^{(\ell)}_{2}\in H^{l}}d(h^{(l)}_{1}, h^{(l)}_{2}) ,\sup_{h^{(\ell)}_{2 \in H^{\ell}}} \inf_{h^{(\ell)}_{1\in H^{l}}}d(h^{(\ell)}_{1}, h^{(\ell)}_{2})\Big\}
\end{equation}
\fi

\paragraph{Capsule Encoded Representations}
To encode paired images $\langle x_1, x_2 \rangle$ into vector pairs $\langle h_1, h_2 \rangle$ the pose vector of each capsule is vectorized and passed as input to a fully connected layer containing 20 activation units. Hence, for each input there is a lower 20-dimensional representation of 32 capsule pose vectors resulting in 512 input features. To ensure all capsules stay active the dropout probability rate is learned for each capsule. The sigmoid function learns the dropout rate of the final capsule layer using \textit{Concrete Dropout} ~\cite{gal2017concrete}, which builds on prior work ~\cite{kingma2015variational, molchanov2017variational} by using a continuous relaxation that approximates the discrete Bernoulli distribution used for dropout, referred to as a concrete distribution. Equation \ref{eq:concrete} shows the objective function for updating the concrete distribution. For a given capsule probability $p_{c}$ in the last capsule layer, the $\mathtt{sigmoid}$ computes the relaxation $\tilde{z}$ on the Bernoulli variable $z$, where $u$ is drawn uniformly between [0,1] where $t$ denotes the temperature values ($t=0.1$ in our experiments) which forces probabilities at the extremum when small. The pathwise derivative estimator is used to find a continuous estimation of the dropout mask.

\begin{equation}\label{eq:concrete}
	\tilde{z}_{t} = \sigma\Big(\frac{1}{t}(\log p_{c} - \log(1-p_{c})) + \log u_{c} - \log(1-u_{c})\Big)
\end{equation}

\iffalse. Gaussian noise is added to the last layer in proportion to the variance of the probability outputs i.e capsules with high probability get less noise added to their pose matrix in the encoding as shown in Equation \ref{eq:noise_encoding}.
\begin{equation}\label{eq:noise_encoding}
	a_{1} = \sigma(\mathcal{M} + b + \epsilon) \quad s.t  \quad \epsilon \sim \mathcal{N}(0, \sigma^{2}_{c^{L}}) 
\end{equation}
\fi

\iffalse
\begin{equation}
	h = \sum_{c}^{C} vec(c_{1,c}) \frown vec(c) 
\end{equation}
\fi

\paragraph{Loss Functions}
% MARGIN LOSS - ORIGINAL DYNAMIC ROUTING PAPER
The original capsule paper with dynamic routing ~\cite{sabour2017dynamic} used a margin loss $L_{c} = T_{c} \max(0, m^{+} - ||v_{c}||)^{2} + \lambda (1 - T_{c})\max(0, ||v_{c}|| -  m^{-})^{2}$ where the class capsule $v_{c}$ has margin $m^{+}=0.9$ positives and $m^{-}=1-m^{+}$ negatives. The weight $\lambda$ is used to prevent the activity vector lengths from deteriorating early in training if a class capsule is absent. The overall loss is then simply the sum of the capsule losses $\sum_{c}L_{c}$. A spread loss~\cite{hinton2018matrix} has also been used to maximize the inter-class distance between the target class and the remaining classes for classifying on the smallNORB dataset. This is given as $L_i = (\max(0, m - (a_t - a_i))^{2}, \quad L = \sum_{i \neq t} L_{i}$ where the margin $m$ is increased linearly during training to ensure lower-level capsule stay active throughout training. This work instead uses a contrastive margin loss ~\cite{chopra2005learning} where the aforementioned capsule encoding similarity function $d_\omega$ outputs a predicted similarity score. \iffalse This distance is $\ell_2$-normalized which ensures the distance is bounded which means the margin is fixed constant during training.\fi The contrastive loss $\mathcal{L}_c$ ensures similar vectorized pose encodings are drawn together and dissimilar poses repulse. Equation \ref{eq:contrastive_loss} shows a a pair of images that are passed to the SCN model where $D_{w} = ||f^{\omega}(x_{1}) - f^{\omega}(x_{2})||^2_{2}$ computes the Euclidean distance between encodings and $m$ is the margin. When using Manhattan distance $D_{w} = \exp\big(-||f^{\omega}(x_{1}) - f^{\omega}(x_{2})||_{1}\big)$ in which case $m \in [0,1)$. \iffalse However, when using the cosine similarity a squared hinge loss $\mathcal{L}(\theta) = \sum_{i=1}^{m}\big(\max(0,1-(\cos(h_{(i,1)},h_{(i,2)})y_{i}))\big)^{2}$\fi \iffalse$0.5 (\sum_{i=1}^{m}(y_{i} - \sigma(w)))$\fi is used where $y \in [-1,1]$. 

\iffalse
By starting with a small margin of 0.2 and linearly increasing it during training to 0.9, we avoid dead capsules in the earlier layers. Spread loss is equivalent to squared Hinge loss with m = 1. Guermeur \& Monfrini (2011) studies a variant of this loss in the context of multi class SVMs. 
\fi

% \sqrt{(h^{(l)}_{1}-h^{(l)}_{2})^{2}}
% D_w was the original \hat{y}_{i} 
\begin{equation}\label{eq:contrastive_loss}
	L_{c}(\omega) = \sum_{i=1}^{m} \Big( \frac{1}{2}(1 - y^{(i)})D^{(i)}_{\omega} + \frac{1}{2}y^{(i)}max(0, m - D^{(i)}_{\omega}) \Big)
\end{equation}

A double margin loss that has been used in prior work by ~\cite{lin2015deephash} is also considered to affect matching pairs such that to account for positive pairs that can also have high variance in the distance measure. It is worth noting this double margin is similar to the aforementioned margin loss used on class capsules, without the use of $\lambda$. Equation \ref{eq:double_margin} shows the double-margin contrastive loss where positive margin $m_p$ and negative margin $m_n$ are used to find better separation between matching and non-matching pairs. This loss is only used for LFW, given the limited number of instances in AT\&T we find the amount of overlap between pairs to be less severe in experimentation. 

\begin{equation}\label{eq:double_margin}
	L_{c}(\omega) = \sum_{i=1}^{m} \big((1 - y^{(i)})max(0, D^{(i)}_{\omega}-m_{n})^{2} + y^{(i)}max(m_{p} - D^{(i)}_{\omega},0)^{2} \big)
\end{equation}

The original reconstruction loss $L_{r}(\theta) = \sum_{j=1}^{2}\sum_{i=1}^{m}(y_{i}^{(j)} - \hat{y}_{i}^{(j)})^{2}$ used as regularization is not used in the pairwise learning setting, instead we rely on the dropout for regularization with exception of the SCN model that uses concrete dropout on the final layer. 

\begin{figure}
\begin{center}
 \includegraphics[scale=0.5]{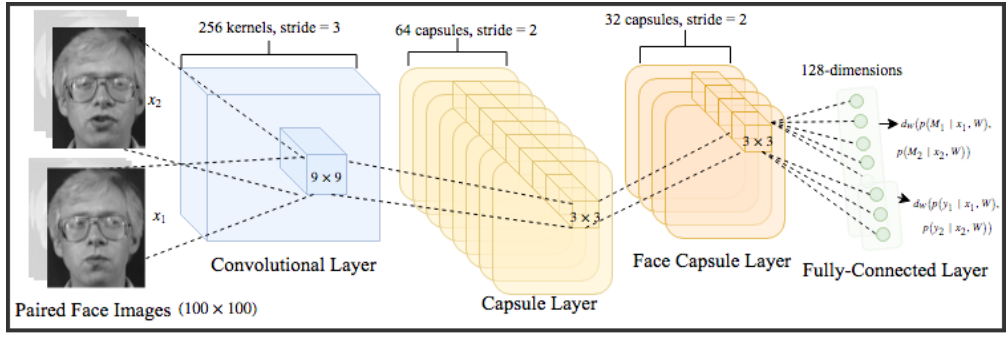}
 \caption{\textit{Siamese Capsule Network Architecture}}\label{fig:scn}
\end{center}
\end{figure}

\iffalse
\begin{equation}\label{eq:contrastive_loss}
	L_{r}(\theta) = \sum_{j=1}^{2}\sum_{i=1}^{m}(y_{i}^{(j)} - \hat{y}_{i}^{(j)})^{2} 
\end{equation}
\fi

\iffalse
\begin{equation}\label{eq:siamese_loss}
	\mathcal{L(W, \theta)} = L_{c}(\theta) +  L_{r}(W)
\end{equation}
\fi

\paragraph{Optimization} Convergence can often be relatively slow for face verification tasks, where few informative batch updates (e.g a sample with significantly different pose for a given class) get large updates but soon after the effect is diminished through gradient exponential averaging (originally introduced to prevent $\alpha \to 0$). Motivated by recent findings that improve adaptive learning rates we use AMSGrad ~\cite{reddi2018convergence}. AMSGrad improves over ADAM in some cases by replacing the exponential average of squared gradients with a maximum that mitigates the issue by keeping long-term memory of past gradients. Thus, AMSGrad does not increase or decrease the learning rate based on gradient changes, avoiding divergent or vanishing step sizes over time. Equation \ref{eq:amsgrad} presents the update rule, where diagonal of gradient $g_t$ is given as $v_t = \Theta_{2}v_{t-1} + (1-\Theta_{2})g_{t}^{2}$, $m_t = \Theta_{1}m_{t-1}+(1-\Theta_{1})g_{t}$, $\alpha_{t} = 1/\sqrt{t}$, $\hat{v}_{t} = \max(\hat{v}_{t-1}, v_{t})$, ensuring $\alpha$ is monotonic.  
 
\begin{equation}\label{eq:amsgrad}
	\omega_{t+1} = \omega_{t} - \alpha \frac{m_{t}}{\sqrt{\hat{v}_{t}+\epsilon}}
\end{equation}

\section{Experiments on Face Verification}
%\paragraph{Datasets} 
% ALL DATASETS - https://www.kairos.com/blog/60-facial-recognition-databases
% http://vis-www.cs.umass.edu/deep_funnel.html
\paragraph{A. AT\&T dataset} The AT\&T face recognition and verification dataset consists of 40 different subjects with only 10 gray-pixel images per subject in a controlled setting. This smaller dataset allows us to test how SCNs perform with little data. For testing, we hold out 5 subjects so that we are testing on unseen subjects, as opposed to training on a given viewpoint of a subject and testing on another viewpoint of the same subject. Hence, zero-shot pairwise prediction is performed during testing.

\paragraph{B. Labeled Faces In The Wild (LFW) dataset} The LFW consists of 13,000 colored photographed faces from the web. This dataset is significantly more complex not only because there 1680 subjects, with some subjects only consisting of two images, but also because of varied amount of aging, pose, gender, lighting and other such natural characteristics. Each image is $250 \times 250$, in this work the image is resized to $100 \times 100$ and normalized. From the original LFW dataset there has been 2 different versions of the dataset that align the images using funneling  ~\cite{huang2007labeled} and deep funneling ~\cite{Huang2012a}. The latter learns to align the images using Restricted Boltzmann Machines with a group sparsity penalty, showing performance improvements for face verification tasks. The penalty leads to an arrangement of the filters that improved the alignment results. This overcomes the problems previous CNNs and models alike had in accounting for pose, orientation and problems Capsule Networks look to address. In contrast, we use the original raw image dataset. 

\iffalse deng2017fine,
   title={Fine-grained face verification: FGLFW database, baselines, and human-DCMN partnership \fi

Both allow for a suitable variety as the former only contains grey-pixel images, a smaller dataset with very few instances per class and images taken in a constrained setting allowing for a more refined analysis, while the LFW data samples are colored images, relatively large with unbalanced classes and taken in an unconstrained setting. 
\iffalse The only constraint on these faces is that they were detected by the Viola-Jones face detector. More details can be found in the technical report below.\fi

% http://vis-www.cs.umass.edu/deep_funnel.html

%~\cite{LFWTechUpdate}

\paragraph{Baselines} SCNs are compared against well-established architectures for image recognition and verification tasks, namely AlexNet, ResNet-34 and InceptionV3 with 6 inception layers instead of the original network that uses 8 layers which are used many of the aforementioned papers in Section \ref{sec:siamese_net}.

\subsection{Results}
Table \ref{tab:results} shows best test results obtained when using contrastive loss with Euclidean distance between encodings (i.e Mahalanobis distance) for both $AT\&T$ and LFW over 100 epochs. The former uses $m=2.0$ and the latter uses $m=0.2$, while for the double margin contrastive loss $m_n = 0.2$ matching margin and  $m_p = 0.5$ negative matching margin is selected. These settings were chosen during 5-fold cross validation, grid searching over possible margin settings. SCN outperforms baselines on the $AT\&T$ dataset after training for 100 epochs. We find that because AT\&T contains far fewer instances an adapted dropout rate leads to a slight increase in contrastive loss. Additionally, adding a reconstruction loss with $\lambda_r=1e^{-4}$ for both paired images led to a decrease in performance when compared to using dropout with a rate $p=0.2$ on all layers except the final layer that encodes the pose vectors. We find for the LFW dataset that the SCN and AlexNet have obtained the best results while SCN has 25\% less parameters. Additionally, the use of a double margin results in better results for the standard SCN but a slight drop in performance when used with concrete dropout on the final layer (i.e SDropCapNet).

\begin{table}
\centering

\begin{tabular}{ccccccc}
 \toprule
 
 & AT\&T & & \multicolumn{2}{c}{LFW \iffalse($10^{-2}$)\fi} & \multicolumn{2}{c}{LFW+Double-M}\iffalse($10^{-2}$)\fi \\
\cmidrule(lr){1-7}Models & Train &Test & Train & Test & Train & Test \\ 
% & CL  & CL & CL & CL \\

  \midrule
 Standard & 0.013 & 0.042 & 0.0021 & 0.012 & 0.0049 & 0.014 \\
 ResNet-34 & 0.015 &0.057 & 0.0018 & 0.012 & 0.0026 &  0.013\\
 AlexNet & 0.032 & 0.085 & \cellcolor{black!20}0.0019 & \cellcolor{black!20}0.009 & \cellcolor{black!20}0.0021 & \cellcolor{black!20}0.010\\
 SCNet & \cellcolor{black!20}0.008 & \cellcolor{black!20}0.019 & 0.0020 & 0.013 & 0.0019 & 0.011  \\
 SDropCapNet &0.010 & 0.032 & \cellcolor{black!20}0.0023 & \cellcolor{black!20}0.010 & 0.0028 & 0.012\\

  \bottomrule
  \end{tabular}  
   \caption{5-fold CV Train \& Test  Contrastive Loss w/ Malahaobonis distance}
  \label{tab:results}
\end{table}

\begin{figure}[!bp]
\centering
\begin{minipage}{.5\textwidth}
  \centering
  \includegraphics[width=1.05\linewidth]{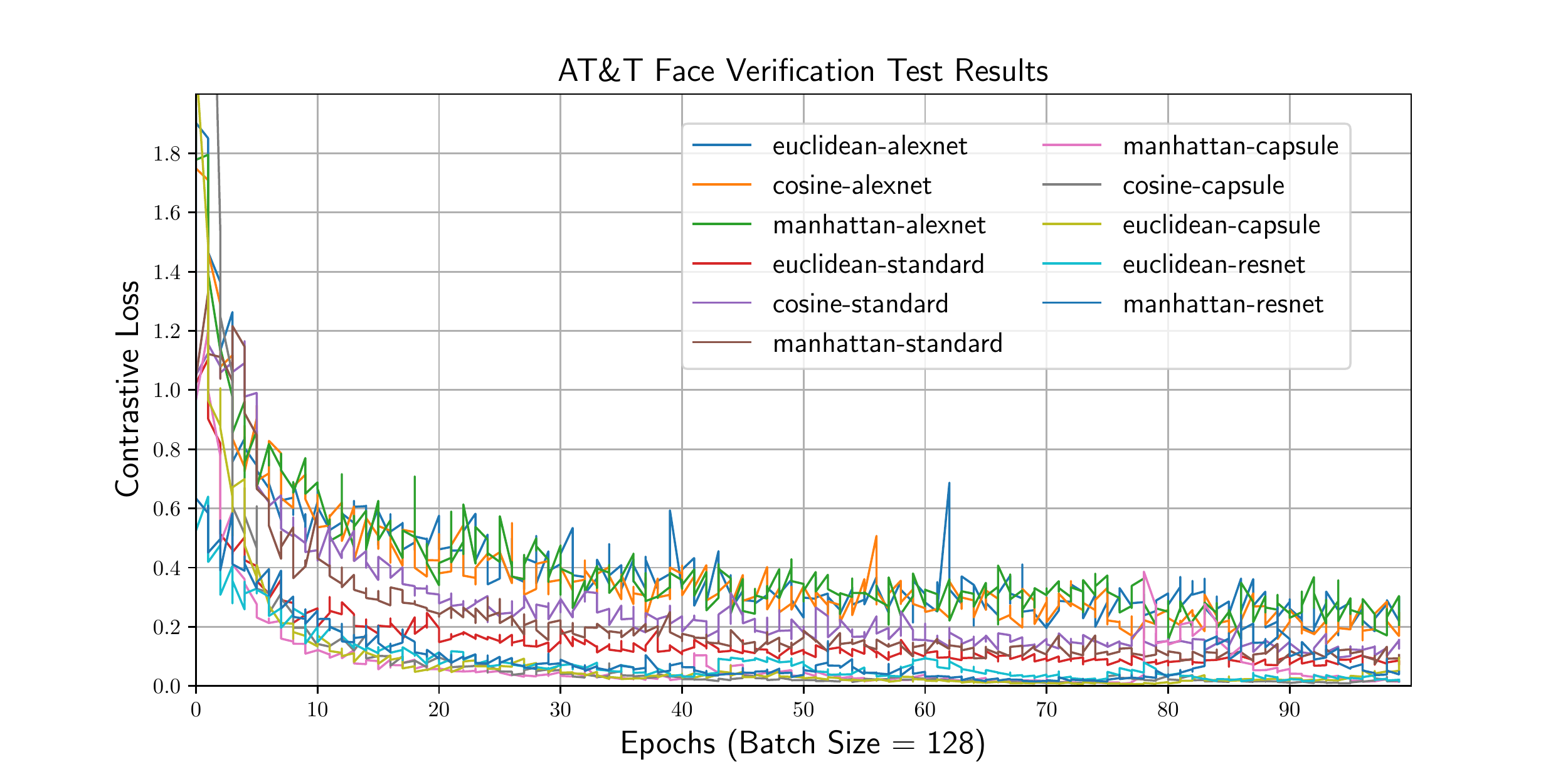}
    \captionof{figure}{\textit{AT\&T Test Set Contrastive Loss}}\label{fig:att_loss}
\end{minipage}%
\begin{minipage}{.5\textwidth}
  \centering
  \includegraphics[width=1.05\linewidth]{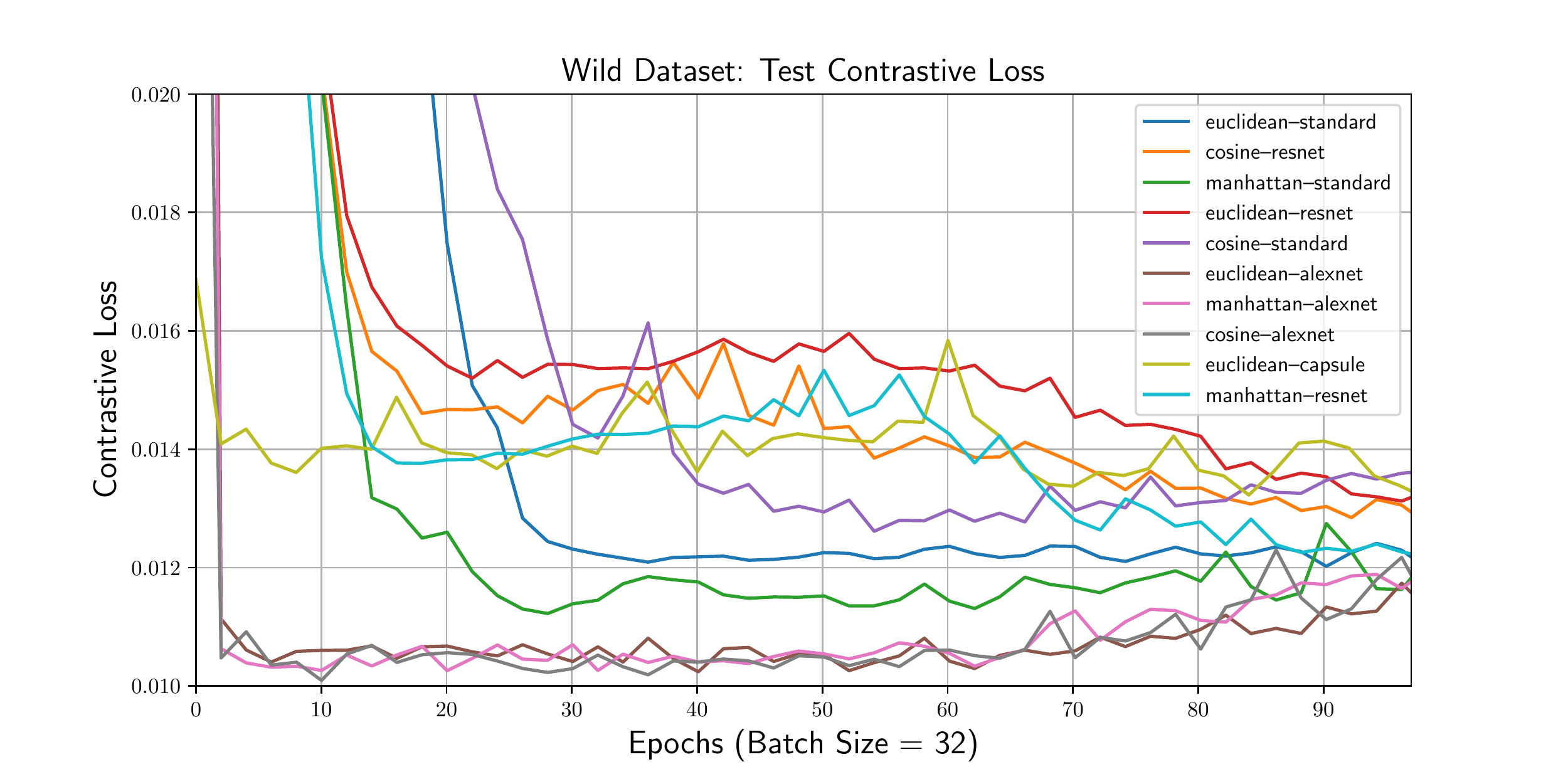}
  \captionof{figure}{\textit{LFW Test Set Contrastive Loss}}\label{fig:lfw_loss}
  \label{fig:test2}
\end{minipage}
\end{figure}

Figure \ref{fig:att_loss} illustrates the contrastive loss during training $\ell_2$-normalized features for each model tested with various distance measures on AT\&T. We find that SCN yields faster convergence on AT\&T, particularly when using Manhattan distance. However for Euclidean distance, we observe a loss variance reduction during training and the best overall performance. Through experiments we find that batch normalized convolutional layers improves performance of the SCN. In batch normalization, $\hat{x}^{(k)} = (x^{(k)} - \mathbb{E}[x^{k}])/\sqrt{Var[x^{(k)}]}$ provides a unit Gaussian batch that is  shifted by $\gamma^{(k)}$ and scaled with $\beta^{(k)}$ so that $a^{(k)} = \gamma^{(k)}\hat{x}^{(k)}+\beta^{(k)}$. This allows the network to learn whether the input range should be more or less diffuse. Batch normalization on the initial convolutional layers reduced variance in loss during training on both the $AT\&T$ and $LFW$ datasets. 
LFW test results shown in in Figure \ref{fig:lfw_loss} show that the SCN model takes longer to converge particularly in the early stages of training, in comparison to AlexNet.

\begin{figure}
\begin{center}
 \includegraphics[scale=0.6]{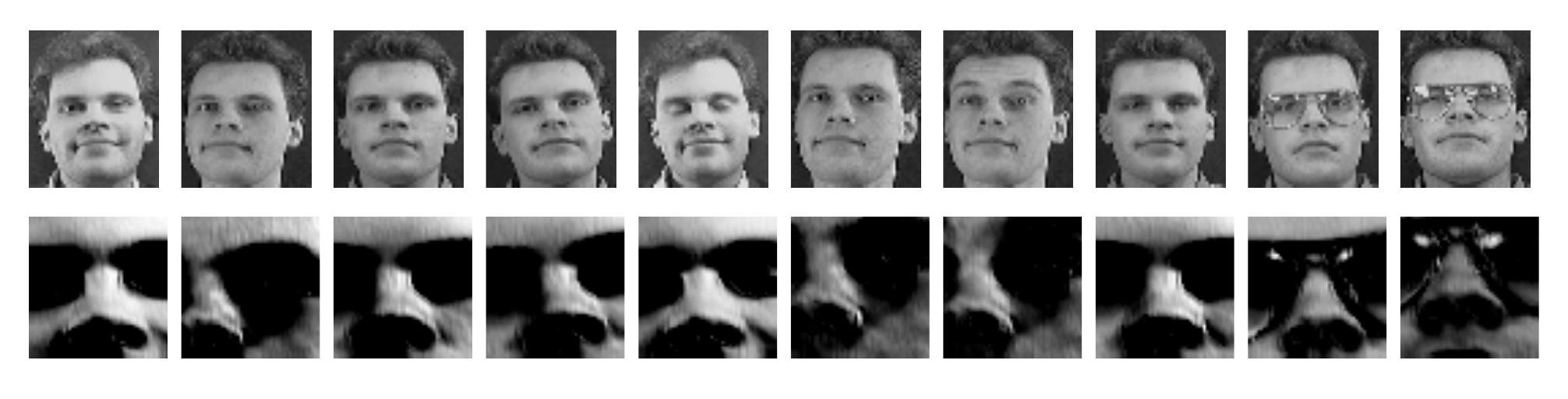}
 \caption{\textit{ResNet Convolutional Layer For AT\&T Face Sample}}\label{fig:att_faces}
\end{center}
\end{figure}

\iffalse
% old version: images/AT_T_Face_Verification_Test_Results.png
\begin{figure}
\begin{center}
 \includegraphics[scale=0.4]
 {images/AT_T_Face_Verification_Test_Results.pdf}
 \caption{\textit{Face Verification Test Set Contrastive Loss} (100 epochs)}\label{fig:loss}
\end{center}
\end{figure}
\fi
To demonstrate some of the difficulties a CNN-based architecture has even with slight change in face orientation, we visualize the lower-level features from the ResNet network as shown in Figure \ref{fig:att_faces}. These are decoded features obtained from larger kernels from the $1^{st}$ convolutional layer. We see that even in the controlled setting, minor rotation of the face from images 2, 6 and 7 (left to right) show that the output is darkened with distorted rotation near the bridge of the nose. This is reflected in the test loss discrepancy when compared with \textit{Siamese Capsule Network}.   

\begin{figure}
\begin{center}
 \includegraphics[scale=0.54]{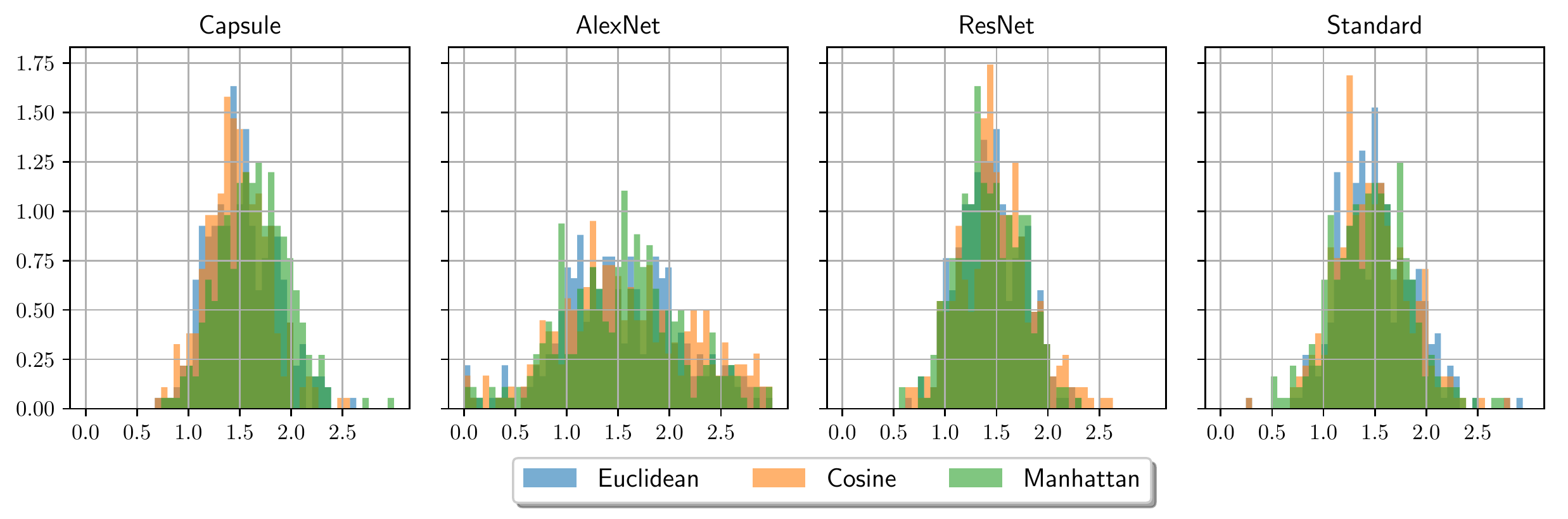}
 \caption{\textit{Probability Density of LFW Positive Pair Test Predictions}}\label{fig:pdf}
\end{center}
\end{figure}

Figure \ref{fig:pdf} shows the probability density of the positive pair predictions for each model for all distances between encodings with contrastive loss for the LFW dataset. We find the variance of predictions is lower in comparison to the remaining models, showing a higher precision in the predictions, particularly for Manhattan distance. Additionally, varying distances for these matching images were close in variance to non-matching images. This motivated the use of the double margin loss considered for the LFW dataset.

Finally, the SCN model has between 104-116 \% less parameters than Alexnet, 24-27 \% Resnet-34 and 127-135\% less than the best standard baseline for both datasets. However, even considering tied weights between models in the SCN, Capsule Networks are primarily limited in speed even with a reduction in parameters due to the routing iterations that are necessary during training.

\section{Conclusion}\label{sec:conclusion}
This paper has introduced the \textit{Siamese Capsule Network}, a novel architecture that extends Capsule Networks to the pairwise learning setting with a feature $\ell_2$-normalized contrastive loss that maximizes inter-class variance and minimizes intra-class variance. The results indicate Capsule Networks perform better at learning from only few examples and converge faster when a contrastive loss is used that takes face embeddings in the form of encoded capsule pose vectors. We find \textit{Siamese Capsule Networks} to perform particularly well on the AT\&T dataset in the few-shot learning setting, which is tested on unseen classes (i.e subjects) during testing, while competitive against baselines for the larger \textit{Labeled Faces In The Wild} dataset.\iffalse Although, the model is slower to train due to the routing iterations for each update.\fi

\iffalse Although \textit{Siamese Capsule Networks} have been demonstrated for face verification we look to utilize these networks for various other pairwise learning tasks.\fi

\iffalse Further improvements are made by using individual dropout rates for the final layer capsule so that more important instantiation parameters are encoded in the last fully-connected layer.\fi \iffalse The model is also significantly smaller by using iterative weight pruning on convolutional layers while retaining the same performance and requiring less parameters than the baseline models.\fi   

\iffalse
\section{Future Work}
Use multiple simultaneous digits and get rid of softmax. Redistribute votes that do not fit the pose of a capsule to other capsules (explainin away). Unsupervised learning of the primary capsules and deeper hierarchies.
How do we get from pixels to the first level parts that output explicit pose parameters ? We would like to do this withoput requiring any labeled data. This \say{de-rendering} stage has to be very nonlinear.
\fi

\bibliography{nips_2018}

\begin{thebibliography}{}
\expandafter\ifx\csname natexlab\endcsname\relax\def\natexlab#1{#1}\fi

\bibitem[{Bahadori(2018)}]{bahadori2018spectral}
Mohammad~Taha Bahadori. 2018.
\newblock Spectral capsule networks .

\bibitem[{Ballard(1987)}]{ballard1987generalizing}
Dana~H Ballard. 1987.
\newblock Generalizing the hough transform to detect arbitrary shapes.
\newblock In {\em Readings in computer vision\/}, Elsevier, pages 714--725.

\bibitem[{Chopra et~al.(2005)Chopra, Hadsell, and LeCun}]{chopra2005learning}
Sumit Chopra, Raia Hadsell, and Yann LeCun. 2005.
\newblock Learning a similarity metric discriminatively, with application to
  face verification.
\newblock In {\em Computer Vision and Pattern Recognition, 2005. CVPR 2005.
  IEEE Computer Society Conference on\/}. IEEE, volume~1, pages 539--546.

\bibitem[{Fischler and Bolles(1987)}]{fischler1987random}
Martin~A Fischler and Robert~C Bolles. 1987.
\newblock Random sample consensus: a paradigm for model fitting with
  applications to image analysis and automated cartography.
\newblock In {\em Readings in computer vision\/}, Elsevier, pages 726--740.

\bibitem[{Gal et~al.(2017)Gal, Hron, and Kendall}]{gal2017concrete}
Yarin Gal, Jiri Hron, and Alex Kendall. 2017.
\newblock Concrete dropout.
\newblock In {\em Advances in Neural Information Processing Systems\/}. pages
  3584--3593.

\bibitem[{Hinton et~al.(2018)Hinton, Sabour, and Frosst}]{hinton2018matrix}
Geoffrey Hinton, Sara Sabour, and Nicholas Frosst. 2018.
\newblock Matrix capsules with em routing.

\bibitem[{Hinton et~al.(2011)Hinton, Krizhevsky, and
  Wang}]{hinton2011transforming}
Geoffrey~E Hinton, Alex Krizhevsky, and Sida~D Wang. 2011.
\newblock Transforming auto-encoders.
\newblock In {\em International Conference on Artificial Neural Networks\/}.
  Springer, pages 44--51.

\bibitem[{Hu et~al.(2014)Hu, Lu, and Tan}]{hu2014discriminative}
Junlin Hu, Jiwen Lu, and Yap-Peng Tan. 2014.
\newblock Discriminative deep metric learning for face verification in the
  wild.
\newblock In {\em Proceedings of the IEEE Conference on Computer Vision and
  Pattern Recognition\/}. pages 1875--1882.

\bibitem[{Huang et~al.(2012)Huang, Mattar, Lee, and
  Learned-Miller}]{Huang2012a}
Gary~B. Huang, Marwan Mattar, Honglak Lee, and Erik Learned-Miller. 2012.
\newblock Learning to align from scratch.
\newblock In {\em NIPS\/}.

\bibitem[{Huang et~al.(2007)Huang, Ramesh, Berg, and
  Learned-Miller}]{huang2007labeled}
Gary~B Huang, Manu Ramesh, Tamara Berg, and Erik Learned-Miller. 2007.
\newblock Labeled faces in the wild: A database for studying face recognition
  in unconstrained environments.
\newblock Technical report, Technical Report 07-49, University of
  Massachusetts, Amherst.

\bibitem[{Humphreys(1983)}]{humphreys1983reference}
Glyn~W Humphreys. 1983.
\newblock Reference frames and shape perception.
\newblock {\em Cognitive Psychology\/} 15(2):151--196.

\bibitem[{Jaderberg et~al.(2015)Jaderberg, Simonyan, Zisserman
  et~al.}]{jaderberg2015spatial}
Max Jaderberg, Karen Simonyan, Andrew Zisserman, et~al. 2015.
\newblock Spatial transformer networks.
\newblock In {\em Advances in neural information processing systems\/}. pages
  2017--2025.

\bibitem[{Kingma et~al.(2015)Kingma, Salimans, and
  Welling}]{kingma2015variational}
Diederik~P Kingma, Tim Salimans, and Max Welling. 2015.
\newblock Variational dropout and the local reparameterization trick.
\newblock In {\em Advances in Neural Information Processing Systems\/}. pages
  2575--2583.

\bibitem[{LaLonde and Bagci(2018)}]{lalonde2018capsules}
Rodney LaLonde and Ulas Bagci. 2018.
\newblock Capsules for object segmentation.
\newblock {\em arXiv preprint arXiv:1804.04241\/} .

\bibitem[{Lin et~al.(2015)Lin, Morere, Chandrasekhar, Veillard, and
  Goh}]{lin2015deephash}
Jie Lin, Olivier Morere, Vijay Chandrasekhar, Antoine Veillard, and Hanlin Goh.
  2015.
\newblock Deephash: Getting regularization, depth and fine-tuning right.
\newblock {\em arXiv preprint arXiv:1501.04711\/} .

\bibitem[{Liu et~al.(2017)Liu, Wen, Yu, Li, Raj, and Song}]{liu2017sphereface}
Weiyang Liu, Yandong Wen, Zhiding Yu, Ming Li, Bhiksha Raj, and Le~Song. 2017.
\newblock Sphereface: Deep hypersphere embedding for face recognition.
\newblock In {\em The IEEE Conference on Computer Vision and Pattern
  Recognition (CVPR)\/}. volume~1.

\bibitem[{McGee(1979)}]{mcgee1979human}
Mark~G McGee. 1979.
\newblock Human spatial abilities: Psychometric studies and environmental,
  genetic, hormonal, and neurological influences.
\newblock {\em Psychological bulletin\/} 86(5):889.

\bibitem[{Molchanov et~al.(2017)Molchanov, Ashukha, and
  Vetrov}]{molchanov2017variational}
Dmitry Molchanov, Arsenii Ashukha, and Dmitry Vetrov. 2017.
\newblock Variational dropout sparsifies deep neural networks.
\newblock {\em arXiv preprint arXiv:1701.05369\/} .

\bibitem[{Ranjan et~al.(2017)Ranjan, Castillo, and Chellappa}]{ranjan2017l2}
Rajeev Ranjan, Carlos~D Castillo, and Rama Chellappa. 2017.
\newblock L2-constrained softmax loss for discriminative face verification.
\newblock {\em arXiv preprint arXiv:1703.09507\/} .

\bibitem[{Reddi et~al.(2018)Reddi, Kale, and Kumar}]{reddi2018convergence}
Sashank~J Reddi, Satyen Kale, and Sanjiv Kumar. 2018.
\newblock On the convergence of adam and beyond.
\newblock In {\em International Conference on Learning Representations\/}.

\bibitem[{Rock(1973)}]{rock1973orientation}
Irvin Rock. 1973.
\newblock {\em Orientation and form\/}.
\newblock Academic Press.

\bibitem[{Sabour et~al.(2017)Sabour, Frosst, and Hinton}]{sabour2017dynamic}
Sara Sabour, Nicholas Frosst, and Geoffrey~E Hinton. 2017.
\newblock Dynamic routing between capsules.
\newblock In {\em Advances in Neural Information Processing Systems\/}. pages
  3859--3869.

\bibitem[{Sankaranarayanan et~al.(2016)Sankaranarayanan, Alavi, and
  Chellappa}]{sankaranarayanan2016triplet}
Swami Sankaranarayanan, Azadeh Alavi, and Rama Chellappa. 2016.
\newblock Triplet similarity embedding for face verification.
\newblock {\em arXiv preprint arXiv:1602.03418\/} .

\bibitem[{Schroff et~al.(2015)Schroff, Kalenichenko, and
  Philbin}]{schroff2015facenet}
Florian Schroff, Dmitry Kalenichenko, and James Philbin. 2015.
\newblock Facenet: A unified embedding for face recognition and clustering.
\newblock In {\em Proceedings of the IEEE conference on computer vision and
  pattern recognition\/}. pages 815--823.

\bibitem[{Sun et~al.(2014)Sun, Chen, Wang, and Tang}]{sun2014deep}
Yi~Sun, Yuheng Chen, Xiaogang Wang, and Xiaoou Tang. 2014.
\newblock Deep learning face representation by joint
  identification-verification.
\newblock In {\em Advances in neural information processing systems\/}. pages
  1988--1996.

\bibitem[{Szegedy et~al.(2015)Szegedy, Liu, Jia, Sermanet, Reed, Anguelov,
  Erhan, Vanhoucke, Rabinovich et~al.}]{szegedy2015going}
Christian Szegedy, Wei Liu, Yangqing Jia, Pierre Sermanet, Scott Reed, Dragomir
  Anguelov, Dumitru Erhan, Vincent Vanhoucke, Andrew Rabinovich, et~al. 2015.
\newblock Going deeper with convolutions.
\newblock Cvpr.

\bibitem[{Taigman et~al.(2014)Taigman, Yang, Ranzato, and
  Wolf}]{taigman2014deepface}
Yaniv Taigman, Ming Yang, Marc'Aurelio Ranzato, and Lior Wolf. 2014.
\newblock Deepface: Closing the gap to human-level performance in face
  verification.
\newblock In {\em Proceedings of the IEEE conference on computer vision and
  pattern recognition\/}. pages 1701--1708.

\bibitem[{Wang and Liu(2018)}]{wang2018optimization}
Dilin Wang and Qiang Liu. 2018.
\newblock An optimization view on dynamic routing between capsules .

\bibitem[{Wen et~al.(2016)Wen, Zhang, Li, and Qiao}]{wen2016discriminative}
Yandong Wen, Kaipeng Zhang, Zhifeng Li, and Yu~Qiao. 2016.
\newblock A discriminative feature learning approach for deep face recognition.
\newblock In {\em European Conference on Computer Vision\/}. Springer, pages
  499--515.

\bibitem[{Zeiler and Fergus(2014)}]{zeiler2014visualizing}
Matthew~D Zeiler and Rob Fergus. 2014.
\newblock Visualizing and understanding convolutional networks.
\newblock In {\em European conference on computer vision\/}. Springer, pages
  818--833.

\end{thebibliography}
\bibliographystyle{nips_2018}

\end{document}